# A Unified Robust Classification Model

**Akiko Takeda, Hiroyuki Mitsugi** TAKEDA@AE.KEIO.AC.JP, KIYUROHI7@Z2.KEIO.JP
Keio University, 3-14-1 Hiyoshi, Kouhoku, Yokohama, Kanagawa 223-8522, Japan

**Takafumi Kanamori** KANAMORI@IS.NAGOYA-U.AC.JP
Nagoya University, Furocho, Chikusa-ku, Nagoya-shi, Aichi 464-8603, Japan

## Abstract

A wide variety of machine learning algorithms such as support vector machine (SVM), minimax probability machine (MPM), and Fisher discriminant analysis (FDA), exist for binary classification. The purpose of this paper is to provide a unified classification model that includes the above models through a robust optimization approach. This unified model has several benefits. One is that the extensions and improvements intended for SVM become applicable to MPM and FDA, and vice versa. Another benefit is to provide theoretical results to above learning methods at once by dealing with the unified model. We give a statistical interpretation of the unified classification model and propose a non-convex optimization algorithm that can be applied to non-convex variants of existing learning methods.

## 1. Introduction

There are a wide variety of machine learning algorithms for binary classification. *Support vector machine* (SVM) is one of the most successful classification algorithms in modern machine learning (Schölkopf & Smola, 2002). The minimax probability machine (MPM) (Lanckriet et al., 2002) and Fisher discriminant analysis (FDA) (Fukunaga, 1990) also address the binary classification problem. Their problem settings assume that only the mean and covariance matrix of each class are known. The optimal hyperplane of MPM is determined by minimizing the worst-case (maximum) probability of misclassification of unseen test samples over all possible class-conditional distributions. FDA is to find a direction which maximizes the projected class means while minimizing the class variance in this direction.

The purpose of this paper is to provide a unified framework for learning algorithms, including SVM, MPM, and FDA, from the viewpoint of robust optimization (Ben-Tal et al., 2009). Robust optimization is an approach that handles optimization problems defined by uncertain inputs. A simple example of robust optimization is

$$\max_{\boldsymbol{w}\in\mathcal{W}}\min_{\boldsymbol{x}\in\mathcal{U}} \boldsymbol{x}^\top\boldsymbol{w}, \tag{1}$$

where $\boldsymbol{w}$ is the parameter to be optimized under the constraint $\boldsymbol{w}\in\mathcal{W}$ and $\boldsymbol{x}$ is an uncertain input in the problem. The *uncertainty set* $\mathcal{U}$ represents the uncertainty of the input. (1) determines the decision making parameter $\boldsymbol{w}$ which maximizes the benefit $\boldsymbol{x}^\top\boldsymbol{w}$ for the worst-case setup among $\boldsymbol{x}\in\mathcal{U}$.

For binary classification, we regard the means $\boldsymbol{x}_+$ and $\boldsymbol{x}_-$ of the data points of each class as uncertain inputs and prepare uncertainty sets $\mathcal{U}_+$ and $\mathcal{U}_-$ of those uncertain inputs. We assume that $\boldsymbol{x}$ of (1) exists in the Minkowski difference $\mathcal{U}$ of $\mathcal{U}_+$ and $\mathcal{U}_-$, i.e.,

$$\mathcal{U}=\mathcal{U}_+\ominus\mathcal{U}_- := \{\boldsymbol{x}_+-\boldsymbol{x}_- \mid \boldsymbol{x}_+\in\mathcal{U}_+,\ \boldsymbol{x}_-\in\mathcal{U}_-\},$$

and define $\mathcal{W}$ by $\{\boldsymbol{w} \mid \|\boldsymbol{w}\|^2=1\}$, where $\|\cdot\|$ is the Euclidean norm. Then we transform (1) into

$$\max_{\boldsymbol{w}:\|\boldsymbol{w}\|^2=1}\ \min_{\boldsymbol{x}_+\in\mathcal{U}_+\boldsymbol{x}_-\in\mathcal{U}_-} (\boldsymbol{x}_+-\boldsymbol{x}_-)^\top\boldsymbol{w}. \tag{2}$$

We call it *robust classification model* (RCM)[1]. This problem always seems to be non-convex because of $\mathcal{W}$. However, it reduces to a convex problem that includes a constraint $\|\boldsymbol{w}\|^2\le 1$ instead of $\|\boldsymbol{w}\|^2=1$ when $\mathcal{U}_+$ and $\mathcal{U}_-$ do not intersect.

[1] Here we used the terminology of "robust" for the model (2) from the notion of "robust optimization", not from the notion of "robust statistics". The aim of the RCM is in providing a unified framework to existing learning methods, not in providing a learning method with better tolerance to outliers.

Appearing in *Proceedings of the 29$^{th}$ International Conference on Machine Learning*, Edinburgh, Scotland, UK, 2012.

In this paper, we show that RCM (2) reduces to the learning methods mentioned above, depending on a prescribed uncertainty set $\mathcal{U}$. For example, we show that MPM is a special case of (2) with an ellipsoidal uncertainty set $\mathcal{U}$. When $\mathcal{U}_+$ and $\mathcal{U}_-$ are defined as reduced convex hulls (Bennett & Bredensteiner, 2000), (2) reduces to $\nu$-SVM (Schölkopf et al., 2000) if $\mathcal{U}_+ \cap \mathcal{U}_- = \emptyset$ and reduces to E$\nu$-SVM (Perez-Cruz et al., 2003), otherwise. The difference between these learning methods turns out only to be in the definition of $\mathcal{U}$ of (2).

The first contribution of handling the unified model (2) is to obtain new learning methods. For example, we can obtain non-convex variants of MPM and FDA by mimicking Perez-Cruz et al.'s extension (Perez-Cruz et al., 2003) from convex $\nu$-SVM to non-convex E$\nu$-SVM.

The second contribution is to provide theoretical results to above learning methods at once by dealing with the unified model (2). Indeed, we provide statistical interpretation for (2) on the basis of the conventional statistical learning theory. We show that (2) with some corresponding uncertainty set is a good approximation for the worst-case minimization of expected loss functions under uncertain probabilities.

We also provide a generalized local optimum search algorithm, that is applicable to non-convex variants of learning models. We prove theoretical results on the local optimum search algorithm.

The paper is organized as follows. In Section 2, we elucidate the unified model, RCM (2), for classification problems. In Section 3, we show RCM's connection with existing learning algorithms and obtain non-convex variants for MPM and FDA in the same way as non-convex E$\nu$-SVM. In Section 4, we give a statistical interpretation of RCM in terms of minimizing the upper and lower bounds of the worst-case expected loss. In Section 5, we describe a local optimum search algorithm for non-convex RCM. We summarize our contributions and future work in Section 6.

# 2. Unified Robust Classification Model

## 2.1. Problem Settings

We shall start by introducing the problem setting and the notations. The observed training samples are denoted as $(\boldsymbol{x}_i, y_i) \in \mathbb{R}^d \times \{+1, -1\}, i \in M := \{1, ..., m\}$. Let $M_+$ be the set of indices of training samples with the label $+1$; likewise for $M_-$. Let $|M_+| = m_+$ and $|M_-| = m_-$, where $|\cdot|$ shows the size of the set.

The goal of the classification task is to obtain a classifier that minimizes the prediction error rate for unseen test samples. For the sake of simplicity, we shall focus on linear classifiers, i.e., $\boldsymbol{x}^\top \boldsymbol{w} + b$ where $\boldsymbol{w}$ ($\in \mathbb{R}^d$) is a vector and $b$ ($\in \mathbb{R}$) is a bias parameter. Most of the discussions in this paper can be directly applied to kernel classifiers (Schölkopf & Smola, 2002). Concretely, the change from $\boldsymbol{x} \in \mathcal{X}$ to the kernel function $k(\cdot, \boldsymbol{x})$ makes statements of Sections 2-4 hold for kernel classifiers, while the algorithm in Section 5 needs small modification.

We shall assume that the training samples are not reliable because of noise or measurement errors. To make a classification model less sensitive to noise in the training samples, we shall focus on representative points of each class, denoted by $\boldsymbol{x}_+$ and $\boldsymbol{x}_-$. These points are not necessarily individual samples, but may be means of the data points of each class. Since the training samples are not reliable, it is reasonable to assume that $\boldsymbol{x}_+$ and $\boldsymbol{x}_-$ will involve some uncertainty. The largest possible sets of $\boldsymbol{x}_+$ and $\boldsymbol{x}_-$ are denoted by $\mathcal{U}_+$ and $\mathcal{U}_-$, respectively, and these sets are defined on the basis of training samples. Throughout this paper, we will assume that both $\mathcal{U}_+$ and $\mathcal{U}_-$ are convex and compact and that they have interior points. Then, their Minkowski difference $\mathcal{U}$ is convex and has a nonempty interior.

The way of constructing the uncertainty set $\mathcal{U}_\pm$ is a very important issue in practice. If we set $\mathcal{U}$ too large in (2), the optimal decision is very robust to uncertain data $\boldsymbol{x}$ but too conservative. Moreover, if we define $\mathcal{U}$ with complicated functions, we cannot easily solve (2). Many robust optimization studies have used polyhedral sets and ellipsoidal sets as $\mathcal{U}$ for the sake of computational tractability. We show examples of $\mathcal{U}_+$ and $\mathcal{U}_-$ in Section 3. We might possibly deal with more complicated problem setting beyond convex $\mathcal{U}_+$ and $\mathcal{U}_-$ by using kernelization techniques.

## 2.2. Properties of RCM

To geometrically interpret RCM (2), Figure 1 shows the ellipsoidal uncertainty sets $\mathcal{U}_+$, $\mathcal{U}_-$ and their Minkowski difference. We can separate the problem (2) into two cases, i.e., whether $\mathcal{U}_+$ and $\mathcal{U}_-$ have an intersection or not, which is equivalent to whether $\mathcal{U}$ includes $\mathbf{0}$ or not. As shown in Theorem 2.2, there is a large difference in computational effort between the two cases. Before giving an intuitive geometric interpretation of RCM in Theorem 2.2, we introduce Lemma 2.1 that further separates the case $\mathbf{0} \in \mathcal{U}$ into two cases: $\mathcal{U}$ includes $\mathbf{0}$ in its interior, $\mathrm{int}(\mathcal{U})$, or on its boundary, $\mathrm{bd}(\mathcal{U})$. In the geometric sense, $\mathbf{0} \notin \mathcal{U}$ holds when $\mathcal{U}_+$ and $\mathcal{U}_-$ are disjoint. $\mathbf{0} \in \mathcal{U}$ implies that $\mathcal{U}_+$ and $\mathcal{U}_-$ are joint. In particular, $\mathbf{0} \in \mathrm{bd}(\mathcal{U})$ implies that $\mathcal{U}_+$ and $\mathcal{U}_-$ touch externally.

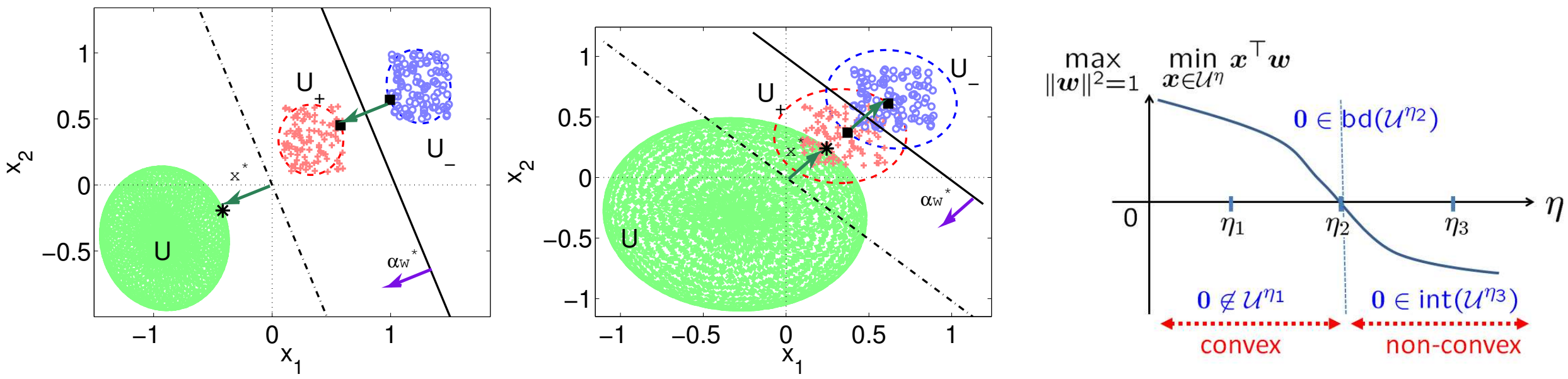

*Figure 1.* Geometric interpretation of RCM (2). **Left:** Two ellipsoidal uncertainty sets, $\mathcal{U}_+$ and $\mathcal{U}_-$, are disjoint ($\mathbf{0} \notin \mathcal{U}$). **Middle:** $\mathcal{U}_+$ and $\mathcal{U}_-$ are joint ($\mathbf{0} \in \mathcal{U}$). The asterisk shows the optimal point $\boldsymbol{x}^*$ in $\mathcal{U}$, and the dash-dot line depicts the hyperplane $\boldsymbol{x}^{*\top}\boldsymbol{w}^* = 0$. The squares are the solutions $\boldsymbol{x}_+^*$ and $\boldsymbol{x}_-^*$ of the inner-minimization on $\mathcal{U}_+ \times \mathcal{U}_-$, and the solid line stands for the optimal hyperplane, $(\boldsymbol{x}_+^* - \boldsymbol{x}_-^*)^\top \boldsymbol{w}^* + b = 0$. The bias term $b$ in the decision function is defined such that the decision boundary passes through the mid-point of the squares. The green arrows indicate the optimal solution $\boldsymbol{x}^*$, and the purple arrows indicate the normal direction $\alpha\boldsymbol{w}^*$ of the hyperplane for some positive $\alpha$. **Right:** Optimal value of RCM (2) with uncertainty set $\mathcal{U}^\eta$.

**Lemma 2.1.** *The optimal value of RCM (2) is positive if and only if $\mathbf{0} \notin \mathcal{U}$. It is zero if and only if $\mathbf{0} \in \mathrm{bd}(\mathcal{U})$, and it is negative if and only if $\mathbf{0} \in \mathrm{int}(\mathcal{U})$.*

We can prove "if" parts by using the supporting hyperplane theorem to three cases ($\mathbf{0} \notin \mathcal{U}$, $\mathbf{0} \in \mathrm{bd}(\mathcal{U})$ and $\mathbf{0} \in \mathrm{int}(\mathcal{U})$). By taking the contrapositive of all the "if" parts, we also can prove "only if" parts.

Let $\mathcal{U}^\eta$ be a parametrized uncertainty set for RCM (2) such that $\mathcal{U}^{\eta_1} \subset \mathcal{U}^{\eta_2}$ holds for $\eta_1 \le \eta_2$. Then the following inequality holds:

$$\max_{\boldsymbol{w}:\|\boldsymbol{w}\|^2=1} \min_{\boldsymbol{x}\in\mathcal{U}^{\eta_1}} \boldsymbol{x}^\top\boldsymbol{w} \ge \max_{\boldsymbol{w}:\|\boldsymbol{w}\|^2=1} \min_{\boldsymbol{x}\in\mathcal{U}^{\eta_2}} \boldsymbol{x}^\top\boldsymbol{w}.$$

This indicates that the optimal value of (2) is non-increasing with respect to the inclusion relation of uncertainty sets. Figure 1 (right) plots the non-increasing optimal value of RCM (2) with respect to $\eta$. An uncertainty set $\mathcal{U}^{\eta_2}$ might exist such that the optimal value of (2) becomes zero.

The following theorem shows that when $\mathbf{0} \notin \mathcal{U}$, the equality constraint $\|\boldsymbol{w}\|^2 = 1$ in (2) can be replaced by $\|\boldsymbol{w}\|^2 \le 1$ without changing the solution. Moreover, $\|\boldsymbol{w}\|^2 = 1$ can be replaced by $\|\boldsymbol{w}\|^2 \ge 1$ when $\mathbf{0} \in \mathcal{U}$. Figure 1 (left and middle) illustrates Theorem 2.2.

**Theorem 2.2.** *For an uncertainty set such that $\mathbf{0} \notin \mathcal{U}$, RCM (2) is equivalent to*

$$\max_{\boldsymbol{w}:\|\boldsymbol{w}\|^2\le 1} \min_{\boldsymbol{x}\in\mathcal{U}} \boldsymbol{x}^\top\boldsymbol{w}. \tag{3}$$

*Moreover, the problem is equivalent to*

$$\min_{\boldsymbol{x}_\pm\in\mathcal{U}_\pm} \|\boldsymbol{x}_+ - \boldsymbol{x}_-\|, \quad \textit{or equivalently,} \quad \min_{\boldsymbol{x}\in\mathcal{U}} \|\boldsymbol{x}\|. \tag{4}$$

*An optimal $\boldsymbol{w}$ of (3) can be obtained from $\boldsymbol{x}^*/\|\boldsymbol{x}^*\|$ by using the optimal $\boldsymbol{x}^* \in \mathcal{U}$ of (4). For an uncertainty set such that $\mathbf{0} \in \mathrm{int}(\mathcal{U})$, RCM (2) is equivalent to*

$$\max_{\boldsymbol{w}:\|\boldsymbol{w}\|^2\ge 1} \min_{\boldsymbol{x}\in\mathcal{U}} \boldsymbol{x}^\top\boldsymbol{w}. \tag{5}$$

*Moreover, the problem is equivalent to $\min_{\boldsymbol{x}\in\overline{\mathcal{U}^c}} \|\boldsymbol{x}\|$, where $\overline{\mathcal{U}^c}$ is the closure of the complement of the convex set $\mathcal{U}$. An optimal $\boldsymbol{w}$ of (5) can be obtained from $-\boldsymbol{x}^*/\|\boldsymbol{x}^*\|$ by using the optimal $\boldsymbol{x}^* \in \overline{\mathcal{U}^c}$.*

*Proof.* Assume $\mathbf{0} \notin \mathcal{U}$. By applying the discussion on the minimum norm duality (Luenberger, 1969) to (3), we can confirm the equivalence of (3) and $\min_{\boldsymbol{x}\in\mathcal{U}} \|\boldsymbol{x}\|$, and the optimal solution $\boldsymbol{w}^* = \boldsymbol{x}^*/\|\boldsymbol{x}^*\|$. On the other hand, in the case of $\mathbf{0} \in \mathrm{int}(\mathcal{U})$, the equivalence of (5) and $\min_{\boldsymbol{x}\in\overline{\mathcal{U}^c}} \|\boldsymbol{x}\|$ is proved from Proposition 3.1 of (Briec, 1997) under the assumption that a convex $\mathcal{U}$ has a nonempty interior. Hence, it is enough to show that there exists an optimal solution $\boldsymbol{w}^*$ of (3) (or (5)) such that $\|\boldsymbol{w}^*\| = 1$, because the difference between (2) and (3) (or (5)) is only the norm constraint of $\boldsymbol{w}$.

Lemma 2.1 ensures that the optimal value of (3) is positive, because

$$\max_{\boldsymbol{w}:\|\boldsymbol{w}\|^2\le 1} \min_{\boldsymbol{x}\in\mathcal{U}} \boldsymbol{x}^\top\boldsymbol{w} \ge \max_{\boldsymbol{w}:\|\boldsymbol{w}\|^2=1} \min_{\boldsymbol{x}\in\mathcal{U}} \boldsymbol{x}^\top\boldsymbol{w} > 0.$$

Since the optimal solution $\boldsymbol{w}^*$ of (3) satisfies $0 < \|\boldsymbol{w}^*\| \le 1$, the following inequalities hold:

$$0 < \min_{\boldsymbol{x}\in\mathcal{U}} \boldsymbol{x}^\top\boldsymbol{w}^* \le \min_{\boldsymbol{x}\in\mathcal{U}} \boldsymbol{x}^\top\boldsymbol{w}^*/\|\boldsymbol{w}^*\| \le \min_{\boldsymbol{x}\in\mathcal{U}} \boldsymbol{x}^\top\boldsymbol{w}^*.$$

The last inequality comes from the optimality of $\boldsymbol{w}^*$. These inequalities imply that $\boldsymbol{w}^*/\|\boldsymbol{w}^*\|$ is also an optimal solution of (3) and that $\|\boldsymbol{w}^*\| = 1$. For the case of $\mathbf{0} \in \mathrm{int}(\mathcal{U})$, we can similarly show that the optimal

*Table 1.* Correspondence with existing classifiers

| $\mathcal{U}$ | $\mathbf{0} \in \mathrm{int}(\mathcal{U})$ | $\mathbf{0} \in \mathrm{bd}(\mathcal{U})$ | $\mathbf{0} \notin \mathcal{U}$ |
|---|---|---|---|
| Ellip. (11) | $\surd$ | MPM | MM-MPM |
| Ellip. (15) | $\surd$ | FDA | FS-FD |
| RCH (8) | E$\nu$-SVM | $\nu_{\min}$ | $\nu$-SVM |
| CH (6) | $\times$ | $\times$ | HM-SVM |

value of (5) is negative and that an optimal solution $\boldsymbol{w}^*$ of (5) exists such that $\|\boldsymbol{w}^*\| = 1$. □

For $\mathbf{0} \in \mathrm{int}(\mathcal{U})$, RCM (2) is essentially a non-convex problem, and we need to use non-convex optimization methods to solve it. Section 5 describes an optimization algorithm for non-convex problems of (2).

## 3. Equivalence to Existing Classifiers

We will show that RCM can be reduced to support vector machine (SVM), minimax probability machine (MPM), or Fisher discriminant analysis (FDA) depending on the prescribed uncertainty set $\mathcal{U}$. In Table 1, "$\times$" means that the corresponding cases never happen. "$\surd$" means that there are no corresponding existing models as far as we know. The models indicated by $\surd$ are the target in this paper.

We denote an optimal solution of (2) as $\boldsymbol{w}^*$ and define the bias term $b$ such that the decision boundary passes through the mid-point of $\boldsymbol{x}_+^*$ and $\boldsymbol{x}_-^*$, i.e., $b = -(\boldsymbol{x}_+^* + \boldsymbol{x}_-^*)^\top \boldsymbol{w}^*/2$. Here, $\boldsymbol{x}_+^* \in \mathcal{U}_+$ and $\boldsymbol{x}_-^* \in \mathcal{U}_-$ stand for the optimal solutions of the inner-minimization in (2) for $\boldsymbol{w} = \boldsymbol{w}^*$.

### 3.1. Hard-Margin SVM, $\nu$-SVM and E$\nu$-SVM

Whenever a data set is linearly separable, there are many hyperplanes that correctly classify all training samples. Vapnik-Chervonenkis theory indicates that a large margin classifier has a small generalization error. The problem can be transformed into a quadratic programming problem and the classification method is called hard-margin support vector classification machine (HM-SVM). Here, we define the uncertainty set (convex hull, CH) as follows:

$$\mathcal{U}_\pm = \mathrm{conv}\{\boldsymbol{x}_i \mid i \in M_\pm\}, \tag{6}$$

where conv means convex hull. By using the Wolfe duality, the equivalence of HM-SVM and RCM (4) is obvious for $\mathcal{U}_+ \cap \mathcal{U}_- = \emptyset$.

HM-SVM has been extended to cope with non-separable data. $C$-SVM (Cortes & Vapnik, 1995) and $\nu$-SVM (Schölkopf et al., 2000) are typical examples of "soft-margin" SVMs. There is a correspondence between $C$-SVM and $\nu$-SVM. That is, the classifier estimated by $C$-SVM with $C \in (0, \infty)$ can be obtained from $\nu$-SVM with a parameter $\nu \in (\nu_{\min}, \nu_{\max}] \subset [0, 1]$, and vice versa. Crisp and Burges (2000) showed $\nu_{\max} = 2\min\{m_+, m_-\}/m$ and gave a geometric interpretation for $\nu_{\min}$. For $\nu \in (\nu_{\max}, 1]$, the optimization problem of $\nu$-SVM is unbounded, and for $\nu \in [0, \nu_{\min})$, $\nu$-SVM provides a trivial solution ($\boldsymbol{w} = \mathbf{0}$ and $b = 0$). Perez-Cruz et al. (2003) devised extended $\nu$-SVM (E$\nu$-SVM) as a way of avoiding such a trivial solution:

$$\begin{aligned} \min_{\boldsymbol{w}, b, \boldsymbol{\xi}, \rho} \; & -\nu\rho + \frac{1}{m}\sum_{i=1}^{m} \xi_i \\ \text{s.t.} \; & y_i(\boldsymbol{x}_i^\top \boldsymbol{w} + b) \geq \rho - \xi_i, \; \xi_i \geq 0, \; i \in M, \; \|\boldsymbol{w}\|^2 = 1. \end{aligned} \tag{7}$$

By forcing the norm of $\boldsymbol{w}$ to be unity, a non-trivial and meaningful solution is obtained for any $\nu \in [0, \nu_{\min})$, but this comes at the expense of convexity. It furthermore provides the same solution as $\nu$-SVM for other values of $\nu$. In that sense, E$\nu$-SVM can be regarded as an extension of $\nu$-SVM. It was experimentally found in (Perez-Cruz et al., 2003) that E$\nu$-SVM often has better generalization performance than $\nu$-SVM.

In order to connect (E)$\nu$-SVM with RCM, we define $\mathcal{U}_\pm^\nu$ as

$$\left\{ \sum_{i \in M_\pm} \lambda_i \boldsymbol{x}_i \Big| \sum_{i \in M_\pm} \lambda_i = 1, \; 0 \leq \lambda_i \leq \frac{2}{\nu m}, \; i \in M_\pm \right\}. \tag{8}$$

The set (8) is essentially equal to a *reduced convex hull* (RCH) (Bennett & Bredensteiner, 2000) or *soft convex hull* (Crisp & Burges, 2000). For linearly non-separable data set, $\mathcal{U}_+^\nu$ and $\mathcal{U}_-^\nu$ intersect with small $\nu$.

Crisp and Burges (2000) showed that $\nu_{\min}$ is the largest $\nu$ such that two RCHs, $\mathcal{U}_+^\nu$ and $\mathcal{U}_-^\nu$, intersect. The model that finds $\nu_{\min}$ corresponds to the case of $\mathbf{0} \in \mathrm{bd}(\mathcal{U}^\nu)$ in the "RCH" of Table 1. Barbero et al. (2012) transformed $\nu$-SVM and E$\nu$-SVM (7) into RCM (2) with $\mathcal{U}_\pm^\nu$ in order to give them a geometric interpretation. Using the results, we can relate $\nu$-SVM, E$\nu$-SVM, and RCM (2) as shown in Table 1.

### 3.2. Minimax Probability Machine and Its Extension

The minimax probability machine (MPM) only uses the mean and covariance matrix of each class for classification tasks (Lanckriet et al., 2002). Suppose that $\boldsymbol{x}_+$ (or $\boldsymbol{x}_-$) is a $d$-dimensional random vector with mean $\bar{\boldsymbol{x}}_+$ (or $\bar{\boldsymbol{x}}_-$) and covariance $\Sigma_+$ (or $\Sigma_-$). We assume that $\bar{\boldsymbol{x}}_+ \neq \bar{\boldsymbol{x}}_-$ and that $\Sigma_\pm$ are positive definite. The MPM minimizes the misclassification probabili-ties under the worst-case setting as follows:

$$\max_{\alpha,\boldsymbol{w},b} \alpha \ \text{ s.t. } \inf_{\boldsymbol{x}_\pm\sim(\bar{\boldsymbol{x}}_\pm,\Sigma_\pm)} \Pr\{\boldsymbol{x}_\pm^\top \boldsymbol{w}+b\geq 0\}\geq\alpha, \quad (9)$$

where $\boldsymbol{x}_+\sim(\bar{\boldsymbol{x}}_+,\Sigma_+)$ refers to the class of distributions that have mean $\bar{\boldsymbol{x}}_+$ and covariance $\Sigma_+$, but are otherwise arbitrary; likewise for $\boldsymbol{x}_-$. In practice, the mean vectors and covariance matrices of each class are estimated from the training samples.

Lanckriet et al. (2002) represented problem (9) as a convex optimization problem known as a second-order cone program (SOCP) and show the dual form:

$$\min_{\kappa} \kappa \quad \text{s.t. } \mathbf{0}\in\mathcal{U}^\kappa := \mathcal{U}_+^\kappa \ominus \mathcal{U}_-^\kappa, \quad (10)$$

$$\text{where} \quad \mathcal{U}_\pm^\kappa = \{\bar{\boldsymbol{x}}_\pm + \Sigma_\pm^{1/2}\boldsymbol{u} \mid \|\boldsymbol{u}\|\leq\kappa\}. \quad (11)$$

$\alpha$ of (9) corresponds to $\kappa$ of (10) as $\kappa=\sqrt{\alpha/(1-\alpha)}$. Therefore, MPM (9) is the problem to find the smallest positive $\kappa$ (denoted by $\kappa_{\max}$) such that the two ellipsoids intersect, i.e., $\mathbf{0}\in\text{bd}(\mathcal{U}^{\kappa_{\max}})$.

The idea of MPM is combined with the idea of the margin maximization in (Nath & Bhattacharyya, 2007). Given acceptable false positive and negative rates, $\eta_+$ and $\eta_-$, the linear classifier can be estimated by

$$\min_{\boldsymbol{w},b} \frac{1}{2}\|\boldsymbol{w}\|^2 \quad \text{s.t.} \sup_{\boldsymbol{x}_\pm\sim(\bar{\boldsymbol{x}}_\pm,\Sigma_\pm)} \Pr\{\boldsymbol{x}_\pm^\top\boldsymbol{w}+b<0\}\leq\eta_\pm. \quad (12)$$

In this paper, we call this model the "margin maximized MPM" (MM-MPM). In the same way as in MPM, (12) can be transformed into an SOCP.

Robust optimization techniques for ellipsoidal uncertainty (Ben-Tal et al., 2009) transform RCM (2) with $\mathcal{U}_\pm=\mathcal{U}_\pm^{\kappa_\pm}$ into

$$\min_{\|\boldsymbol{w}\|^2=1} \kappa_+\|\Sigma_+^{1/2}\boldsymbol{w}\|+\kappa_-\|\Sigma_-^{1/2}\boldsymbol{w}\|-(\bar{\boldsymbol{x}}_+-\bar{\boldsymbol{x}}_-)^\top\boldsymbol{w}. \quad (13)$$

We define $\kappa_+^{\max}$ and $\kappa_-^{\max}$ as constants such that $\mathcal{U}_+^{\kappa_+}$ and $\mathcal{U}_-^{\kappa_-}$ touch. For $\kappa_\pm\in[0,\kappa_\pm^{\max})$, $\mathcal{U}_+^{\kappa_+}\cap\mathcal{U}_-^{\kappa_-}=\emptyset$ holds, and RCM (13) is equivalent to MM-MPM (12) with $\kappa_\pm=\sqrt{(1-\eta_\pm)/\eta_\pm}$. We can confirm this by comparing the dual form of MM-MPM and the dual of (13), that is equivalent to (4). Furthermore, (13) with $\kappa_\pm=\kappa_{\max}$ coincides with MPM (9) (see Table 1).

### 3.3. Fisher Discriminant Analysis and Its Extension

In Fisher discriminant analysis (FDA) as in MPM (9), a discriminant hyperplane is computed from the means and covariances of random vectors $\boldsymbol{x}_+$ and $\boldsymbol{x}_-$. The hyperplane is determined from the optimal solution $\boldsymbol{w}^*$ to the following problem (Fukunaga, 1990):

$$\max_{\boldsymbol{w}} \frac{(\bar{\boldsymbol{x}}_+-\bar{\boldsymbol{x}}_-)^\top\boldsymbol{w}}{\|(\Sigma_++\Sigma_-)^{1/2}\boldsymbol{w}\|}. \quad (14)$$

The problem finds a direction which maximizes the projected class means while minimizing the class variance in this direction.

Likewise for MPM, FDA has a probabilistic interpretation under the worst-case scenario. Using the ellipsoidal uncertainty set defined by

$$\mathcal{U}^\zeta=\{\boldsymbol{x}=(\bar{\boldsymbol{x}}_+-\bar{\boldsymbol{x}}_-)+(\Sigma_++\Sigma_-)^{1/2}\boldsymbol{u} \mid \|\boldsymbol{u}\|\leq\zeta\}, \quad (15)$$

FDA (14) can be represented as

$$\min_{\zeta} \zeta \quad \text{s.t. } \mathbf{0}\in\mathcal{U}^\zeta. \quad (16)$$

FDA can be extended to RCM (2) with the uncertainty set $\mathcal{U}^\zeta$ for a prescribed parameter $\zeta>0$. Let $\zeta_{\max}$ be the optimal value of (16). Then, along the same lines as the MPM in Section 3.2, we find that RCM (2) with $\mathcal{U}=\mathcal{U}^{\zeta_{\max}}$ is equivalent to FDA.

Indeed, RCM (2) with $\mathcal{U}^\zeta$ is transformed into

$$\min_{\|\boldsymbol{w}\|^2=1} \zeta\|(\Sigma_++\Sigma_-)^{1/2}\boldsymbol{w}\|-(\bar{\boldsymbol{x}}_+-\bar{\boldsymbol{x}}_-)^\top\boldsymbol{w}.$$

Especially for $\zeta\in[0,\zeta_{\max})$, the norm constraint is replaced with the convex constraint $\|\boldsymbol{w}\|^2\leq 1$ without changing the optimal solution. Here, MM-FDA refers to this estimator. In replacing the Euclidean norm $\|\boldsymbol{w}\|$ with the $L_1$-norm $\|\boldsymbol{w}\|_1$, MM-FDA is equivalent to a sparse feature selection model based on FDA (FS-FD) (Bhattacharyya, 2004).

## 4. Statistical Interpretation for RCM

We can give a statistical interpretation for RCM on the basis of statistical learning theory. Let us start by introducing a loss function $\ell:\mathbb{R}\to\mathbb{R}$ that defines the loss of the decision function $\boldsymbol{x}^\top\boldsymbol{w}+b$ regarding the sample $(\boldsymbol{x},y)$ as $\ell(y(\boldsymbol{x}^\top\boldsymbol{w}+b))$.

A goal of the classification task is to obtain an accurate classifier. For this purpose, it is reasonable to minimize the expected loss, $\mathbb{E}[\ell(y(\boldsymbol{x}^\top\boldsymbol{w}+b))]$, with respect to $\boldsymbol{w}$ and $b$. Let us define $p(\boldsymbol{x}|y)$ as the conditional probability density of $\boldsymbol{x}$, given the binary label $y$, and $\pi_+$ and $\pi_-$ as the marginal probabilities of the positive and negative labels, respectively. $\mathbb{E}[\ell(y(\boldsymbol{x}^\top\boldsymbol{w}+b))]$ is computed by

$$\begin{aligned}&\pi_+\int\ell(\boldsymbol{x}^\top\boldsymbol{w}+b)p(\boldsymbol{x}|+1)d\boldsymbol{x}\\ &+\pi_-\int\ell(-(\boldsymbol{x}^\top\boldsymbol{w}+b))p(\boldsymbol{x}|-1)d\boldsymbol{x}.\end{aligned}$$

Since the true probability distribution is unknown, we cannot minimize the expected loss directly.

Now let us consider the ambiguity of the probability distribution $p(\boldsymbol{x}|y)$. Let $\mathcal{P}_+$ and $\mathcal{P}_-$ be sets of probability densities. Each set of probabilities expresses the uncertainty of the conditional probabilities $p(\boldsymbol{x}|+1)$ and $p(\boldsymbol{x}|-1)$, respectively. We can use the min-max decision rule for the uncertainty of $p(\boldsymbol{x}|y)$ as follows:

$$\min_{\boldsymbol{w}:\|\boldsymbol{w}\|^2=1} \max_{p(\boldsymbol{x}|\pm 1)\in\mathcal{P}_\pm} \min_{b\in\mathbb{R}} \mathbb{E}[\ell(y(\boldsymbol{x}^\top \boldsymbol{w}+b))]. \quad (17)$$

The worst-case minimization problem is difficult to solve. Therefore, we propose to solve RCM (2), since we can prove that RCM (2) is a good approximation for minimizing the worst-case expected loss.

To relate (17) and RCM, we firstly give an equivalent formulation for RCM. Here, we define $\boldsymbol{x}_+$ and $\boldsymbol{x}_-$ as the mean of the input vector $\boldsymbol{x}$ under the conditional probabilities $p(\boldsymbol{x}|+1)$ and $p(\boldsymbol{x}|-1)$, respectively, i.e., $\boldsymbol{x}_\pm = \int \boldsymbol{x} p(\boldsymbol{x}|\pm 1) d\boldsymbol{x}$. Here, we assume that all probability distributions in $\mathcal{P}_\pm$ have the mean vector. Let $\mathcal{U}_+$ and $\mathcal{U}_-$ be

$$\mathcal{U}_\pm = \left\{ \int \boldsymbol{x} p(\boldsymbol{x}|\pm 1) d\boldsymbol{x} \;\middle|\; p(\boldsymbol{x}|\pm 1) \in \mathcal{P}_\pm \right\}. \quad (18)$$

Suppose that the uncertainty sets of probability densities, $\mathcal{P}_\pm$, are both convex; i.e., a mixture of two probability densities also lies in the uncertainty set. Then $\mathcal{U}_+$ and $\mathcal{U}_-$ are convex sets.

**Theorem 4.1.** *Suppose that $\ell(z)$ is a non-increasing function. An optimal solution of the RCM with the uncertainty sets $\mathcal{U}_+$ and $\mathcal{U}_-$ in (18) is also optimal to*

$$\min_{\boldsymbol{w}:\|\boldsymbol{w}\|^2=1} \max_{\boldsymbol{x}_\pm\in\mathcal{U}_\pm} \min_{b\in\mathbb{R}} J_\ell(\boldsymbol{w}, b; \boldsymbol{x}_+, \boldsymbol{x}_-), \quad (19)$$

*where*

$$J_\ell(\boldsymbol{w}, b; \boldsymbol{x}_+, \boldsymbol{x}_-) = \pi_+ \ell(\boldsymbol{x}_+^\top \boldsymbol{w} + b) + \pi_- \ell(-\boldsymbol{x}_-^\top \boldsymbol{w} - b).$$

*Proof.* For a fixed $\boldsymbol{w}$ and $\boldsymbol{x}_\pm \in \mathcal{U}_\pm$, minimizing $J_\ell(\boldsymbol{w}, b; \boldsymbol{x}_+, \boldsymbol{x}_-)$ respect to $b$ is equivalent to

$$\min_{b'} \pi_+ \ell((\boldsymbol{x}_+ - \boldsymbol{x}_-)^\top \boldsymbol{w} - b') + \pi_- \ell(b').$$

Since the objective function above is non-increasing in $(\boldsymbol{x}_+ - \boldsymbol{x}_-)^\top \boldsymbol{w}$, there exists a non-increasing function $\phi(z)$ such that

$$\phi((\boldsymbol{x}_+ - \boldsymbol{x}_-)^\top \boldsymbol{w}) = \min_b J_\ell(\boldsymbol{w}, b; \boldsymbol{x}_+, \boldsymbol{x}_-).$$

Hence, one has

$$\begin{aligned} &\min_{\boldsymbol{w}:\|\boldsymbol{w}\|^2=1} \max_{\boldsymbol{x}_\pm\in\mathcal{U}_\pm} \min_{b} \; J_\ell(\boldsymbol{w}, b; \boldsymbol{x}_+, \boldsymbol{x}_-) \\ &= \phi( \max_{\boldsymbol{w}:\|\boldsymbol{w}\|^2=1} \min_{\boldsymbol{x}_\pm\in\mathcal{U}_\pm} (\boldsymbol{x}_+ - \boldsymbol{x}_-)^\top \boldsymbol{w}). \end{aligned}$$

As a result, the optimal solution of the RCM is also optimal for problem (19). □

**Theorem 4.2.** *We assume that i) $\ell(z)$ is convex, decreasing, and second-order differentiable, and that ii) $0 \le \ell''(z) \le L \in \mathbb{R}$ holds for all $z$. Suppose that $\boldsymbol{x}_\pm \in \mathcal{U}_\pm$ is in the ball with the radius $c$, i.e., $\|\boldsymbol{x}_\pm\| \le c$. Then, for the optimal value $J^*$ of (19), one has*

$$\min_{\substack{\boldsymbol{w}: \\ \|\boldsymbol{w}\|^2=1}} \max_{p(\boldsymbol{x}|\pm 1)\in\mathcal{P}_\pm} \min_{b\in\mathbb{R}} \mathbb{E}[\ell(y(\boldsymbol{x}^\top \boldsymbol{w}+b))] \in [J^*, J^* + \frac{Lc^2}{2}].$$

*Proof.* The convexity of $\ell(z)$ leads to a lower bound, $J_\ell(\boldsymbol{w}, b; \boldsymbol{x}_+, \boldsymbol{x}_-)$, and the Taylor expansions of $\ell(z)$ around $z = \boldsymbol{x}_+^\top \boldsymbol{w} + b$ and $z = \boldsymbol{x}_-^\top \boldsymbol{w} + b$ yield an upper bound, $J_\ell(\boldsymbol{w}, b; \boldsymbol{x}_+, \boldsymbol{x}_-) + \frac{Lc^2}{2}$, of $\mathbb{E}[\ell(y(\boldsymbol{x}^\top \boldsymbol{w} + b))]$. Even when the min-max operation is applied, $J^*$ and $J^* + \frac{Lc^2}{2}$ remain bounds for the worst-case expected loss (17). □

The theorem implies that problem (19) minimizes the bounds of (17). Noticing that the optimal solution of problem (19) is available by solving RCM as shown in Theorem 4.1, Theorem 4.2 implies that RCM minimizes the upper and lower bounds of the worst-case expected loss (17) at the same time.

There are various ways to estimate the bias term $b$ for RCM. The simplest way is to use $b^* = -(\boldsymbol{x}_+^* + \boldsymbol{x}_-^*)^\top \boldsymbol{w}^*/2$. Another promising method is to construct an appropriate statistical model for the projected samples $(\boldsymbol{x}_i^\top \boldsymbol{w}^*, y_i), i \in M$. The projected samples, $\boldsymbol{x}_i^\top \boldsymbol{w}^*$, $i \in M$, are scattered in one-dimensional space, from which we can estimate $b$ on the basis of the statistical model.

## 5. Solution Method for RCM

The RCM has a significantly larger range of parameter $\kappa$ or $\zeta$ than an existing convex model such as MPM, MM-MPM, FDA or FS-FD (see Table 1). Therefore, the RCM enhances a possibility of improving these existing classification models. Indeed, Perez-Cruz et al. (2003) experimentally showed that the generalization performance of E$\nu$-SVM is often better than that of original $\nu$-SVM. In this section, we propose a solution method that is generalized from the local algorithms of (Perez-Cruz et al., 2003; Takeda & Sugiyama, 2008).

### 5.1. Two-stage Optimization Strategy

Suppose that we solve RCM (2) with the uncertainty set $\mathcal{U}^\eta$ with one parameter $\eta$ and that $\mathcal{U}^{\eta_1} \subset \mathrm{int}(\mathcal{U}^{\eta_2})$ holds for $\eta_1 < \eta_2$. Let us define $\eta_{\max}$ such that the optimal value of (2) with $\mathcal{U} = \mathcal{U}^{\eta_{\max}}$ is zero.

First, we need to compute $\eta_{\max}$ in order to confirm that the given problem (2) is essentially convex or not.

**Algorithm 5.1.**

**Step 1:** Choose $\widetilde{\boldsymbol{w}}_0$ satisfying $\|\widetilde{\boldsymbol{w}}_0\| = 1$ and $\epsilon > 0$. Let $t \leftarrow 0$.
**Step 2:** Solve the following program:

$$\max_{\boldsymbol{w}} g(\boldsymbol{w}) \quad \text{s.t.} \quad \widetilde{\boldsymbol{w}}_t^\top \boldsymbol{w} = 1, \tag{21}$$

where $g(\boldsymbol{w}) = \min_{\boldsymbol{x}\in\mathcal{U}} \boldsymbol{x}^\top \boldsymbol{w}$, and let the optimal solution be $\widehat{\boldsymbol{w}}_t^*$.
**Step 3:** If $\|\widetilde{\boldsymbol{w}}_t - \widehat{\boldsymbol{w}}_t^*\| \le \epsilon$, terminate and output $\widetilde{\boldsymbol{w}}^* \leftarrow \widetilde{\boldsymbol{w}}_t$.
**Step 4:** Otherwise, let $\widetilde{\boldsymbol{w}}_{t+1} \leftarrow \widehat{\boldsymbol{w}}_t^*/\|\widehat{\boldsymbol{w}}_t^*\|$. Let $t \leftarrow t+1$. Repeat Steps 2–4.

*Figure 2.* Local optimum search algorithm for non-convex RCM with $\eta > \eta_{\max}$, that is, $\mathbf{0} \in \mathrm{int}(\mathcal{U})$.

The parameter $\eta_{\max}$ is obtained as the optimal solution of the convex problem:

$$\min_{\eta} \eta \quad \text{s.t.} \quad \mathbf{0} \in \mathcal{U}^\eta. \tag{20}$$

When $\mathcal{U}_\pm^\eta$ are ellipsoidal sets of (11) (or (15)), the problem reduces to MPM (10) (or FDA (16)). When $\mathcal{U}_\pm^\eta$ are RCHs, the problem reduces to a linear programming problem and gives us $\nu_{\min}$.

If the input parameter $\eta$ is equal to $\eta_{\max}$, we have already obtained an optimal solution from (20). If $\eta < \eta_{\max}$, we next solve the convex problem (4) by using a standard optimization software.

### 5.2. Local Optimization Algorithm for Non-convex RCM

For $\eta > \eta_{\max}$, RCM (2) is essentially equivalent to (5) that includes a non-convex constraint, $\|\boldsymbol{w}\|^2 \ge 1$. We next need to solve (2) as a non-convex problem.

In the area of global optimization, non-convex RCM (5) (precisely, a problem constructed by taking dual for the inner-minimization in (5)) is known as a *reverse convex program* (RCP), or *canonical d.c. programming*. This differs from a conventional convex program only by the presence of a reverse convex constraint ($\|\boldsymbol{w}\|^2 \ge 1$ in the current case). When all functions are linear except for the reverse convex constraint, the RCP problem is especially called *linear reverse convex program* (LRCP). E$\nu$-SVM is an LRCP, for which Perez-Cruz et al. (2003) proposed a local optimum search algorithm and Takeda and Sugiyama (2008) proposed a global optimum search algorithm.

Here, we show a local optimum search algorithm (Algorithm 5.1) that is generalized from the local algorithms (Perez-Cruz et al., 2003; Takeda & Sugiyama, 2008) of E$\nu$-SVM for non-convex RCM. It is essentially the same as the local algorithm, Algorithm 7, in (Takeda & Sugiyama, 2008) when $\mathcal{U}$ of $g(\boldsymbol{w})$ is an RCH (8) and $\epsilon = 0$.

RCM (2) requires maximizing $g(\boldsymbol{w}) = \min_{\boldsymbol{x}\in\mathcal{U}} \boldsymbol{x}^\top \boldsymbol{w}$ subject to a non-convex constraint, $\boldsymbol{w}^\top \boldsymbol{w} = 1$. Instead of solving the non-convex problem directly, we can iteratively solve the relaxation problems (21) (in Algorithm 5.1). Since $g(\boldsymbol{w})$ is concave, (21) can be solved by using convex minimization techniques.

The non-convex constraint of (2) is linearized at $\widetilde{\boldsymbol{w}}_t$ in the algorithm, and the linear constraint $\widetilde{\boldsymbol{w}}_t^\top \boldsymbol{w} = 1$ is updated every iteration. Note that the negativity of the optimal value of (21) is guaranteed because of $\mathbf{0} \in \mathrm{int}(\mathcal{U})$. As the algorithm proceeds, the solutions $\widetilde{\boldsymbol{w}}_t$ improve, i.e.,

$$g(\widetilde{\boldsymbol{w}}_t) \le g(\widehat{\boldsymbol{w}}_t^*) < g(\widehat{\boldsymbol{w}}_t^*/\|\widehat{\boldsymbol{w}}_t^*\|) = g(\widetilde{\boldsymbol{w}}_{t+1}) < 0, \tag{22}$$

because $\widetilde{\boldsymbol{w}}_t^\top \widehat{\boldsymbol{w}}_t^* = 1$ together with $\widetilde{\boldsymbol{w}}_t^\top \widetilde{\boldsymbol{w}}_t = 1$ implies $\|\widehat{\boldsymbol{w}}_t^*\| > 1$. Note that $\widetilde{\boldsymbol{w}}_t$ is a feasible solution for (21). Hence, if (21) has no better solutions than $\widetilde{\boldsymbol{w}}_t$, $\widetilde{\boldsymbol{w}}_t$ is returned as an optimal solution $\widehat{\boldsymbol{w}}_t^*$ of (21). The algorithm terminates after that.

The computation of $g(\boldsymbol{w})$ may be difficult for general uncertainty sets. However, we do not need an explicit formula for $g(\boldsymbol{w})$ in (21). If $\mathcal{U}$ is a convex set, we can obtain a dual formulation (max-problem) for $g(\boldsymbol{w})$ and replace the max-min problem (21) with a simple max-problem, that is, a one-level convex problem. Indeed, when the uncertainty set is an RCH (8) of data points, we can take the dual for $g(\boldsymbol{w}) = \min_{\boldsymbol{x}\in\mathcal{U}} \boldsymbol{x}^\top \boldsymbol{w}$ and change (21) into a linearized E$\nu$-SVM (7) whose constraint is $\widetilde{\boldsymbol{w}}_t^\top \boldsymbol{w} = 1$ instead of $\|\boldsymbol{w}\|^2 = 1$. When the algorithm is applied to the RCM having ellipsoidal uncertainty, we analytically obtain the optimal value $g(\boldsymbol{w})$ for any $\boldsymbol{w}$. Indeed, for ellipsoidal uncertainty (11), $g(\boldsymbol{w})$ is equal to the one derived by multiplying the objective function of (13) by -1.

**Theorem 5.2.** *For any $\epsilon > 0$, Algorithm 5.1 terminates in a finite number of iterations.*

*Proof.* Let the negative value $g_{\mathrm{opt}}$ be the optimal value of RCM (2). Suppose $\|\widehat{\boldsymbol{w}}_t^*\| > 1$ for all $t = 1, 2, \ldots$. Otherwise, the algorithm terminates. By evaluating $g(\widetilde{\boldsymbol{w}}_{t+1}) - g(\widetilde{\boldsymbol{w}}_t)$, we have

$$\sum_{t=0}^{\infty} \left(g(\widetilde{\boldsymbol{w}}_{t+1}) - g(\widetilde{\boldsymbol{w}}_t)\right) \ge \sum_{t=0}^{\infty} \left(\frac{1}{\|\widehat{\boldsymbol{w}}_t^*\|} - 1\right) g_{\mathrm{opt}} > 0.$$

The above inequality and the boundedness of $g(\widetilde{\boldsymbol{w}}_t)$ lead to $\lim_{t\to\infty} \frac{1}{\|\widehat{\boldsymbol{w}}_t^*\|} - 1 = 0$. Therefore, $\gamma_t$ exists such that $\|\widehat{\boldsymbol{w}}_t^*\| = 1 + \gamma_t$, $0 < \gamma_t = o(1)$, that leads

to $\|\widetilde{\boldsymbol{w}}_t - \widehat{\boldsymbol{w}}_t^*\| = \sqrt{2\gamma_t + \gamma_t^2}$. Since $\gamma_t \to 0$ holds, the stopping rule $\|\widetilde{\boldsymbol{w}}_t - \widehat{\boldsymbol{w}}_t^*\| \le \epsilon$ with positive $\epsilon$ is satisfied in a finite number of iterations. □

We can show that Algorithm 5.1 with $\epsilon = 0$ terminates within a finite number of iterations when the uncertainty set of RCM (2) is represented by a convex polyhedron by mimicking the proof of Theorem 8 for E$\nu$-SVM in (Takeda & Sugiyama, 2008).

Here, suppose that $g(\widetilde{\boldsymbol{w}}^*)$ is differentiable, i.e., $g(\boldsymbol{w})$ has a unique subgradient at $\widetilde{\boldsymbol{w}}^*$ as $\partial g(\widetilde{\boldsymbol{w}}^*) = \arg\min_{\boldsymbol{x}\in\mathcal{U}} \boldsymbol{x}^\top \widetilde{\boldsymbol{w}}^* = \{\nabla g(\widetilde{\boldsymbol{w}}^*)\}$. For example, $g(\boldsymbol{w})$ is differentiable under ellipsoidal uncertainty (11) (or (15)). Then Theorem 5.3 shows a sufficient condition for the local optimality of the solution $\widetilde{\boldsymbol{w}}^*$ if it is obtained by Algorithm 5.1 with $\epsilon = 0$.

**Theorem 5.3.** *Suppose that $g(\widetilde{\boldsymbol{w}}^*)$ is differentiable. Algorithm 5.1 that terminates with $\epsilon = 0$ provides a local solution $\widetilde{\boldsymbol{w}}^*$ to RCM (2) when the maximum eigenvalue of $\nabla^2 g(\widetilde{\boldsymbol{w}}^*)$ is less than $g(\widetilde{\boldsymbol{w}}^*)$.*

*Proof.* Note that $\widetilde{\boldsymbol{w}}^*$ is the optimal solution of (21) at the final iteration. Therefore, $\widetilde{\boldsymbol{w}}^*$ satisfies the first- and second-order necessary conditions:

$$\nabla g(\widetilde{\boldsymbol{w}}^*) + \eta \widetilde{\boldsymbol{w}}^* = \boldsymbol{0}, \tag{23}$$

$$\boldsymbol{d}^\top \nabla^2 g(\widetilde{\boldsymbol{w}}^*)\boldsymbol{d} \le 0,\ \forall \boldsymbol{d} \text{ such that } \widetilde{\boldsymbol{w}}^{*\top}\boldsymbol{d} = 0, \tag{24}$$

where $\eta$ is a Lagrange multiplier. We can show $\eta = -g(\widetilde{\boldsymbol{w}}^*) > 0$ by noticing that $\nabla g(\widetilde{\boldsymbol{w}}^*)$ is a minimizer to $\min_{\boldsymbol{x}\in\mathcal{U}} \boldsymbol{x}^\top \widetilde{\boldsymbol{w}}^* = g(\widetilde{\boldsymbol{w}}^*)$ and using (23). When the maximum eigenvalue of $\nabla^2 g(\widetilde{\boldsymbol{w}}^*)$ is less than $-\eta$, $\widetilde{\boldsymbol{w}}^*$ satisfying (23) and (24) also satisfies the second-order sufficient conditions for the local optimality of (2):

$$\nabla g(\boldsymbol{w}) + 2\zeta \boldsymbol{w} = \boldsymbol{0}, \quad \boldsymbol{d}^\top(\nabla^2 g(\boldsymbol{w}) + 2\zeta \boldsymbol{I})\boldsymbol{d} < 0,$$
$$\forall \boldsymbol{d} \neq \boldsymbol{0} \text{ such that } \boldsymbol{w}^\top \boldsymbol{d} = 0,$$

where $\boldsymbol{I}$ is the identity matrix and $\zeta(\ge 0)$ is a multiplier. This shows that $\widetilde{\boldsymbol{w}}^*$ is a local solution of (2). □

Theorem 5.3 may be extendable to the non-differentiable case of $g(\widetilde{\boldsymbol{w}}^*)$, though more assumptions are necessary (see Theorems 3.2.16, 3.2.20 and 3.2.21 in (Polak, 1997)).

## 6. Conclusions

We developed the robust classification model (RCM), a model which includes SVM, MPM and FDA for specific uncertainty sets. The choice of uncertainty set is significant in this model. This model enables extensions and improvements to SVM to be applied to MPM and FDA, and vice versa.

The unified model will be of help in clarifying relationships among existing models and in finding new classifiers and new algorithms. That is, we might be able to devise a new classifier by finding a reasonable uncertainty set for RCM. It will be important to see how the learning algorithm, uncertainty set, and prediction accuracy relate to each other.